\def\year{2021}\relax
\documentclass[letterpaper]{article} 
\usepackage{aaai21}  
\usepackage{times}  
\usepackage{helvet} 
\usepackage{courier}  
\usepackage[hyphens]{url}  
\usepackage{graphicx} 

\usepackage{natbib}  
\usepackage{caption} 

\usepackage[utf8]{inputenc} 
\usepackage{booktabs}       
\usepackage{amsfonts}       
\usepackage{nicefrac}       
\usepackage{microtype}      
\usepackage{pgfplots}
\usepackage{comment}
\usepackage{paralist}
\usepackage{dsfont}
\usepackage[ruled,linesnumbered]{algorithm2e}

\usepackage{pgfplots}
\pgfplotsset{compat=newest}
\definecolor{crimson}{rgb}{0.86, 0.08, 0.24}
\definecolor{carrotorange}{rgb}{0.93, 0.57, 0.13}
\definecolor{goldenyellow}{rgb}{1.0, 0.87, 0.0}
\definecolor{darkspringgreen}{rgb}{0.09, 0.45, 0.27}
\definecolor{denim}{rgb}{0.08, 0.38, 0.74}
\definecolor{blue(ncs)}{rgb}{0.0, 0.53, 0.74}
\definecolor{dandelion}{rgb}{0.94, 0.88, 0.19}
\definecolor{darkspringgreen}{rgb}{0.09, 0.45, 0.27}
\definecolor{amethyst}{rgb}{0.6, 0.4, 0.8}
\definecolor{carnelian}{rgb}{0.7, 0.11, 0.11}
\definecolor{darkbrown}{rgb}{0.4, 0.26, 0.13}

\usetikzlibrary{patterns}
\usepgfplotslibrary{fillbetween}

\usepackage{graphicx}
\usepackage{tikz}

\usepackage[utf8]{inputenc}
\usepackage[english]{babel}

\newtheorem{thm}{Theorem}[subsection]
\newtheorem{lemma}[thm]{Lemma}
\newtheorem{proof}{Proof}[section]

\usepackage{amsmath}
\usepackage{xfrac}    
\usepackage{amsmath}  
\usepackage{nicefrac} 
\newcommand{\rom}[1]{%
  \textup{\uppercase\expandafter{\romannumeral#1}}%
}

\usepackage{booktabs}       
\usepackage{multirow}
\usepackage{longtable}
\usepackage{siunitx}

\usepackage{subcaption}
\usepackage[switch]{lineno}

\usepackage{array}

\title{Learning to Stop: Dynamic Simulation Monte-Carlo Tree Search}
\author {
    Li-Cheng Lan\textsuperscript{\rm 1},
Meng-Yu Tsai\textsuperscript{\rm 2}, 
Ti-Rong Wu\textsuperscript{\rm 2}, 
I-Chen Wu\textsuperscript{\rm 2,3},
Cho-Jui Hsieh\textsuperscript{\rm 1} \\
}
\affiliations{
   \textsuperscript{\rm 1} Department of Computer Science, UCLA, Los Angeles, USA\\
    \textsuperscript{\rm 2} Department of Computer Science, National Chiao-Tung University Taiwan\\
    \textsuperscript{\rm 3} Research Center for IT Innovation, Academia Sinica, Taiwan\\
    lclan@cs.ucla.edu, \{adam0923686343,kds285\}@gmail.com, icwu@csie.nctu.edu.tw, chohsieh@cs.ucla.edu
    
}
\begin{document}
\maketitle

\begin{abstract}

Monte Carlo tree search (MCTS) has achieved state-of-the-art results in many domains such as Go and Atari games when combining with deep neural networks (DNNs). When more simulations are executed, MCTS can achieve higher performance but also requires enormous amounts of CPU and GPU resources. However, not all states require a long searching time to identify the best action that the agent can find. For example, in 19x19 Go and NoGo, we found that for more than half of the states, the best action predicted by DNN remains unchanged even after searching 2 minutes. This implies that a significant amount of resources can be saved if we are able to stop the searching earlier when we are confident with the current searching result. In this paper, we propose to achieve this goal by predicting the uncertainty of the current searching status and use the result to decide whether we should stop searching. With our algorithm, called Dynamic Simulation MCTS (DS-MCTS), we can speed up a NoGo agent trained by AlphaZero 2.5 times faster while maintaining a similar winning rate. Also, under the same average simulation count, our method can achieve 61\% winning rate against the original program.


\end{abstract}

\section{Introduction}
Many of the state-of-the-art AI programs in games such as  Go, Atari, shogi, chess, and NoGo \cite{silver2016mastering,silver2017mastering, silver2018general, schrittwieser2019mastering,  lan2019multiple} use a combination of deep neural networks (DNNs) \cite{lecun2015deep} and Monte Carlo tree search (MCTS) \cite{browne2012survey,kocsis2006bandit}. Starting from 2015, DNNs have been used to extract high-level information from a given state and provide high-quality policy and state value for MCTS agents. Even the raw DNN, without any lookahead, can achieve a strong professional level \cite{silver2017mastering}. The well-trained networks can then used by policy value MCTS (PV-MCTS). PV-MCTS uses the policy as a heuristic to narrow down each node's search space in MCTS, while the state value is used to make each evaluation more accurate.  After using PV-MCTS, AlphaGo Zero can achieve 5,185 Elo, which corresponds to more than 99\% winning rate against the raw DNN.

\begin{figure}[t]
  \centering
   
        \begin{tikzpicture}[font=\small]
            
            \begin{axis} [   
                xlabel = Data Percentage (\%),
                ylabel = Simulation Count,
                grid style=dashed,
                width=8cm,
                height=4cm,
                enlarge x limits={abs=0pt},
                enlarge y limits={abs=0pt},
                ymajorgrids=true,
                yticklabel style = {font=\small},
                xticklabel style = {font=\small},
                every extra y tick/.style={
                    grid=none, 
                    tick0/.initial=black,
                    tick1/.initial=black,
                    tick2/.initial=black,
                    tick3/.initial=black,
                    tick4/.initial=black,
                    yticklabel style={
                     color=\pgfkeysvalueof{/pgfplots/tick\ticknum},
                    },
                ticklabel style = {font=\small}
                },
                extra y ticks ={692},
            ]
            
            \addplot [
            color=black,
            solid,
            line width=0.7pt,
            densely dotted,
            ]
            coordinates{
             (0, 692)(100,692)
            }
            ;

            \addplot [
            name path=A,
            color=black,
            thick,
            ] 
            table[
            col sep=comma, 
            header=true, 
            x=norm_idx,
            y=msc,
            ] 
            {Figures/msc_under6400.csv};
            
            \addplot[name path=line,line width=0pt] coordinates{
             (0,0)(100,0)
            };
            
            \addplot[name path=top_line,line width=0pt] coordinates{
             (0,6400)(100,6400)
            };

            \addplot fill between[
            of = A and line,
            soft clip={domain=-3:0},
            every segment no 0/.style={
            fill=gray,
            fill opacity=0.2,
            %
            }
            ];
            

    \end{axis}
    \end{tikzpicture}

  \caption{The minimum simulation count required to find the best move under a total of 6400 simulations in the 19x19 Go.}
  \label{fig:go19_msc}
\end{figure}
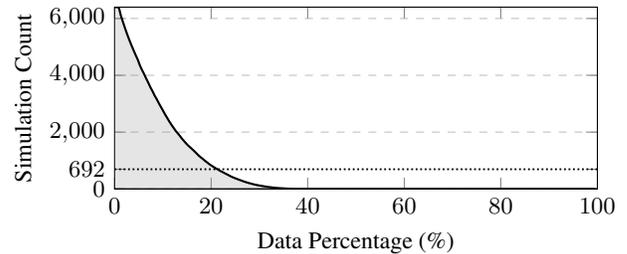

Although achieving remarkable results, PV-MCTS requires enormous amounts of GPU resources in both training and testing. For example, AlphaGo used more than a thousand CPUs and 250 GPUs for playing games against Lee Sedol, not to mention the millions of self-play required during training time for AlphaZero. Many methods have been proposed to reduce the computation of PV-MCTS~\cite{lan2019multiple,gao2018three} or decrease the inference time by parallelization~\cite{ Liu2020Watch}. However, seldom of them consider terminating the search after the stable best action is found. Fig.~\ref{fig:go19_msc} shows the minimum simulation count at which the best action remains the best choice using PV-MCTS with a total of 6400 simulations for each position for 19x19 Go. The results show that more than 62\% of the states only require 1 MCTS simulation, and the average minimum simulation count is only 692. Moreover, KataGo~\cite{wu2019accelerating} speeds up AlphaZero training by randomly reducing the simulations on nearly 75\% of states during self-play. We expect more acceleration if we reduce the simulation of each state carefully.



On the other hand, understanding the certainty of our prediction is critical for both humans and models in many domains \cite{amodei2016concrete, gal2016dropout, guo2017calibration, lakshminarayanan2017simple, liang2017enhancing}. One of the reasons is that we can gather more information or even ask experts to avoid being wrong on highly uncertain data. The uncertainty estimation is even more important in reinforcement learning \cite{clements2019estimating}. Some use it for better exploration \cite{pathak2017curiosity, burda2018exploration}, others use it for avoiding risky or unknown actions 
\cite{garcia2015comprehensive}. However, to the best of our knowledge, non of previous works can be implemented on the tree searching directly since the uncertainty of tree searching is very different from modern neural networks. For example, given a state, we will become certain if we found a winning strategy after searching several simulations. Hence, we can predict the uncertainty based on different search stage.

\begin{figure}
    \centering
    \begin{subfigure}[b]{0.48\linewidth}        
        \centering

        \includegraphics[width=0.7\linewidth]{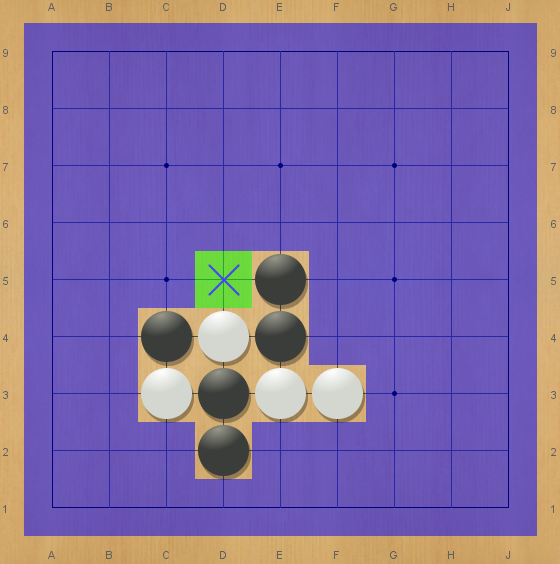}
        \caption{Easy state.}
        \label{fig:easy_state}
    \end{subfigure}
    \begin{subfigure}[b]{0.48\linewidth}        
        \centering
        \includegraphics[width=0.7\linewidth]{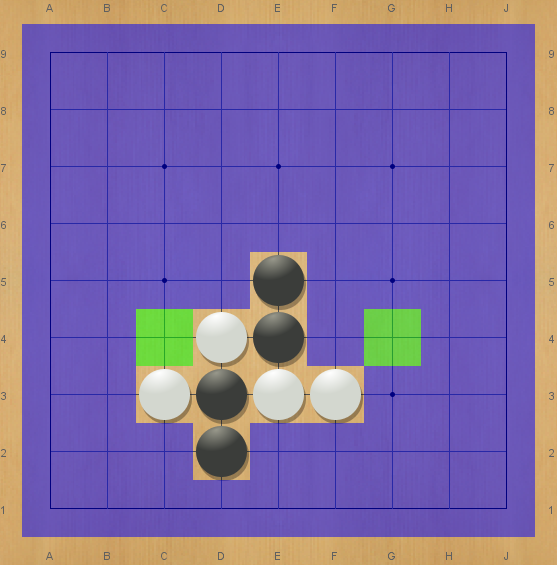}
        \caption{Hard state.}
        \label{fig:hard_state}
    \end{subfigure}
    \caption{The illustration of the state uncertainty in the 9x9 Go. The color shows the policy output by the neural network. Purple through bright green ranges from the lower to the higher probability.}
    \label{fig:easy_hard_state}

\end{figure}

In this paper, we first define the certainty of an MCTS searching status as the confidence of the current best action remains unchanged after reaching the given maximum simulations. Take Fig.~\ref{fig:easy_state}, an easy state of 9x9 Go, as an example. Even beginner Go players know that marked point is the only move to consider; DNN policies can also learn this. Therefore, we should be certain about our decision after evaluating the root state and stop the search immediately. On the contrary, Fig.~\ref{fig:hard_state} shows a rigid state where even an agent trained by the AlphaZero algorithm is uncertain about which action is better after a small amount of searching. 

We proposed three ways to predict the uncertainty. The first method uses the policy of the original neural network and the confidence calibration technique~\cite{guo2017calibration}. The second and the third methods train an auxiliary network to estimate the confidence. The first and second methods only use the given state's information, while the third method also includes information about the current search status. The results show that all the methods can achieve more than 92\% accuracy. 

Finally, we proposed {\it Dynamic Simulation MCTS} (DS-MCTS), which dynamically determines whether to continue the MCTS search according to the current search status's uncertainty. The experiment results on a NoGo AlphaZero agent show that the proposed algorithm leads to a speedup of 2.5 while maintaining a similar winning rate. The result is significant because playing at the level of AlphaZero; even a single lousy move will lead to losing immediately. Moreover, by increasing the maximum simulations, we achieve 61\% win rate against the original PV-MCTS using the same number of average simulations. We also conduct experiments on 9x9 Go. Our method speeds up 1.66x even if we do not fine-tune our hyper-parameters. 
\section{Background}
Policy Value Monte Carlo Tree Search (PV-MCTS), used in the AlphaGo Zero program \cite{silver2017mastering}, is similar to MCTS \cite{browne2012survey,kocsis2006bandit} in that it is a best-first search algorithm that uses Monte Carlo simulation to evaluate state values. It incorporates into MCTS a two-headed neural network, which we refer to as a policy value neural network (PV-NN). Given a state $s$, PV-NN outputs a policy distribution $p(*|s)$ that represents the probability $p(a|s)$ of each action $a$, and a scalar $v(s)$, which estimates the state value.

PV-MCTS contains four phases for each simulation: (1) \emph{Selection}: starting from the root state $s_0$, traverse downwards, choosing the best move until reaching a leaf state $s_T$. At every time step $t$, AlphaZero uses the PUCT algorithm \cite{rosin2011multi} as the tree search policy to select the action for state $s_t$:
\begin{equation}
\label{equ:PUCT}
     a_{t} = \mathop{\arg\max}_{a} \Bigg\{Q(s_t,a)+c_{PUCT}\times p(a|s_t)\times\frac {\sqrt{\sum_{b}{N(s_t,b)}}}{1+N(s_t,a)}\Bigg\},
\end{equation}
where $Q(s,a)$ and $N(s,a)$ represents the estimated value of the search tree and visit count of action $a$ for state $s$, $c_{PUCT}$ is a constant value for exploration, and $P(s,a)$ is the prior probability provided by PV-NN. (2) \emph{Expansion}: Expand the search tree from the selected leaf $s_T$. (3) \emph{Evaluation}: The neural network evaluates the selected leaf state $s_T$ using the PV-NN (4) \emph{Update}: The evaluation result of PV-NN is then used to update the prior probability $P(s_T,*)$ of $s_T$ and the action values $Q(s_t,a_t)$ on the selected path ($s_0, s_1, ..., s_T$). After maximum simulations have been executed, PV-MCTS returns a policy $\pi(s_0,a)=N(s_0,a)^{1/\tau}/\sum_{b}{N(s_0,b)^{1/\tau}}$ of root state $s_0$, where $\tau$ is a temperature parameter that can be adjusted by users. A typical choice is to set $\tau \to 0^+$, which means the move with the maximum visit count is always selected.

\section{Method}
This section will first discuss the proposed uncertainty estimation algorithm for MCTS and then show how to use it to conduct a Dynamic Simulation MCTS (DS-MCTS).

\subsection{Uncertainty estimation}
Evaluating the uncertainties is challenging since the "ground truth" of uncertainty estimates are usually not available. Therefore, we try to define our "uncertainty" of current search status based on the following observations:

\begin{itemize}
\item 
Although there is no "ground truth" for playing games, PV-MCTS will converge to a decent result after hundreds of simulations have been executed. On the other hand, it is much easier to predict the uncertainty when a large enough maximum simulation counts $N_{max}$ exists, since we can use the policy and action values after $N_{max}$ simulations to establish our "ground truth".

\item 
There could be more than one best action (e.g., two possible winning moves), and it is sufficient to find any of these actions. The state should be considered certain for the agent if it is sure that the action it found is nearly the best. The near-best actions can be defined as actions with similar highest action value after $N_{max}$ simulations. 

\item  Even if the agent can predict the best policy at the beginning of the search, a state should be considered uncertain if another sub-optimal move is considered the best action during the search since luck plays a factor in the beginning stages of a search. Action can be considered sub-optimal if its action value is lower than the best action after $N_{max}$ simulations. 

\end{itemize}

Based on the above observations, we propose the following definition of our uncertainty. 

Let $\pi(a|s,n), Q(s,a,n)$ be the policy and the action value provided by the PV-MCTS agent after $n$ simulations on the state $s$. We use  $\tau \to 0^+$ to select $\pi$ (only select the best action). If an action $a'$ hasn't been visited, the action value $Q(s,a',n)$ will be $-1$, which means losing. Given a maximum simulation $N_{max}$, we use the action value $Q(s,a,N_{max})$ to approximate the true action value. Therefore, we have the approximate reward $R$ of current policy as follows:
\begin{equation}
\label{equ:approximate_reward}
    R_{N_{max}}(s,n) = \sum\limits_{a}(\pi(a|s,n)*Q(s,a,N_{max})).
\end{equation}
We define an MCTS tree status after $n$ simulations is uncertain $U(s,n)=1$ if:
\begin{equation}
\label{equ:uncertainty}
    \exists n' \geq n  (R_{N_{max}}(s,N_{max}) - R_{N_{max}}(s,n') \geq \epsilon), 
\end{equation}
where $\epsilon$ is a small constant that we can tolerate. We are certain about the tree status $U(s,n)=0$, when every $n'\geq n$ select one of the near-best actions. This definition satisfies our three observations. First, the uncertainty is based on the given $N_{max}$. Second, it is sufficient to find any of the near-best action within the $\epsilon$ threshold. Finally, the search status after $n$th simulations will be considered uncertain ($U(s,n)=1$) if any sub-optimal action be chosen at any $n' \geq n$. 

For convenience sake, we define the minimum simulation count $M(s)$ of a state $s$ as: after $M(s)$ simulations, the searching status becomes \emph{certain}. In another words, $U(s,n)=1$ if $n<M(s)$, otherwise, $U(s,n)=0$ if $M(s) \leq n \leq N_{max}$. Fig. \ref{fig:go19_msc} is drawn by $M(s)$. The state shown in Fig. \ref{fig:easy_state} has minimum simulation count $M(s_{easy})=1$ and the state of Fig. \ref{fig:hard_state} has a minimum simulation count $M(s_{hard})$ that is close to $N_{max}$.

Our main goal is to stop the search as soon as $M(s)$ simulations have been executed. However, predicting $M(s)$ is hard for neural networks since this requires the models to predict the uncertainty of the future. Hence, we let our models predict the probability of being uncertain $U(s,n)=1$ after $n$ simulation have been executed on state $s$ by providing a value $u$. If $u$ is large, it is more likely $U(s,n)=1$. With the value $u$, we can stop the searching if $u$ is smaller than a tuned threshold.

\subsubsection{Using the original PV-NN}
Although our ultimate goal is predicting the uncertainty based on both the given state and the current tree status, it is still important to predict without any tree status at the beginning of the search since more than half of the states can be certain about its choice once the root state is evaluated. Part of the reason is the high quality of the original PV-NN, and its policy will directly influence the search. According to the PUCT algorithm, if an action's prior probability $p(a|s)$ is large, the selection algorithm of PUCT (Equ.~\ref{equ:PUCT}) will encourage the agent to explore the action, and as a result, it will have a higher chance to end up with the most simulation count and be selected as the best action. Therefore, in this section, we use the original PV-NN to predict the uncertainty at the beginning of the search.

One straightforward way to use the original PV-NN is by using the probability of best action. If the probability of the best action is 0.99, then it is almost certain that it is the best action the agent will choose after the search. However, according to \citet{guo2017calibration}, the softmax prediction of, say, a ResNet~\cite{he2016deep} may not match its true confidence. It needs to be calibrated. For instance,  \citet{guo2017calibration} showed that Platt Scaling \cite{platt1999probabilistic} (temperature scaling) is a simple but surprisingly effective way to calibrate the prediction. It searches for a proper temperature $\tau$ to adjust the prediction $\pi$. For untrustworthy models it will set $\tau \to \infty$, while for trustworthy models  $\tau \to 0^+$. Based on this idea, we approximate the value $u$ with the wrong prediction probability, shown as follows: 
\begin{equation}
\label{equ:CF}
u = 1-\max_a \frac{p(a|s)^{1/\tau}}{\sum\limits_{a'}(p(a'|s)^{1/\tau}}.
\end{equation}
In the experiments, we show that this approach can moderately reduce the average search count, but there is still a large room for improvement. In addition, although using the original PV-NN can provide a usable $u$, it does not match our uncertainty definition. For example, it will be uncertain if the final best action is different from its prediction, even if its prediction is a near-best action. Also, the original PV-NN does not make use of the information of the tree states. We thus propose estimating uncertainty using an auxiliary neural network in the following section.

\subsubsection{Using auxiliary neural networks}
In this subsection, we propose two kinds of auxiliary neural networks to provide the value $u$, which is both trained by predicting our uncertainty $U(s,n)$ directly.
The first auxiliary neural network takes only the state as input and is used at the beginning of the tree search. We call this the state uncertainty network (State-UN). The second auxiliary neural network takes both the state and the current MCTS tree status information as input, which is called the MCTS uncertainty network (MCTS-UN).

\subsubsection{State uncertainty network (State-UN)} 
We generate training data by the following steps. First, to increase the diversity, we generate self-play records using a small simulation count with Dirichlet noise. Second, we re-evaluate the states of those self-play records with a higher simulation count $N_{max}$ without any noise to get the uncertainty $U(s,n)$ of each simulation.

The State-UN uses the same state input features as original PV-NN. The outputs of state Uncertainty Network are $\text{StateUN(s)} = (u,\mathbf{p},v)$, which are used to predict: 1) the beginning uncertainty $U(s,1)$ defined in Equation~\ref{equ:uncertainty}, 2) MCTS policy $\pi(*|s,N_{max})$ after $N_{max}$ simulations, and 3) the final result $z \in \{1,-1\}$ of the game record. We then get the following loss function 
\begin{equation}
    \label{equ: state-un-loss}
    l =  (u - U(s,1))^2 + c_1 ((v-z)^2 - \pmb{\pi}^T log(\mathbf{p}))  + c_2\Vert{\theta}\Vert^2, 
\end{equation}
where $c_1, c_2$ are weighting constants. The first term of the loss function is to make $u$ close to $U(s,1)$ using mean squared error. The second term of loss function, controlled by $c_1$, is the same as AlphaGo Zero \cite{silver2017mastering} training its policy and value. Both $\pmb{\pi}$ and $\mathbf{p}$ are vectors and $\pmb{\pi}^T log(\mathbf{p})$ is their cross entropy. We add this term in order to help the network extract more information from state features. The experiment results shows that with a small $c_1=0.1$ can truly help the performance of $u$ and don't really need to tune $c_1$. The last term is for regularization, where $\theta$ is the parameters of our model.

\subsubsection{MCTS Uncertainty Network (MCTS-UN)} 
As more simulations have been executed, more states will switch from uncertain to certain. It is worthy to evaluate the uncertainty according to the latest search tree status. Hence, we propose MCTS Uncertainty Network (MCTS-UN) method to identify those states using both the state and current MCTS information at any time during the search.

For efficiency, given an $N_{max}$, we hope to train our MCTS-UN on a single simulation $n$ and be used on other simulations $n'$. In this way, we can check the uncertainty using only one model (MCTS-UN) at any time of the search and stop the search as soon as possible. It is critical to select a proper $n$ to ensure the model can be generalized to different simulations. We select the $n$ that minimizes the average MCTS count with only one checkpoint for identifying the uncertain states during the search. We also assume that the accuracy of negative samples ($U(s,n)=0$) will have a limitation $\alpha$ when we make the accuracy of positive samples $U(s,n)=1$ close to $100\%$. This can be done by using a very small threshold $thr$ and only predict negative when the output $u$ is smaller than the small threshold. With this predictor, the minimum of our average simulation count if we perform the uncertainty at $n$ will be:
\begin{equation}
\begin{split}
     f(\alpha,n)=
     (n \times \alpha + N_{max} \times (1-\alpha)) p(M(s) < n)\\
     + N_{max}p(M(s)\geq n)
\end{split}
\end{equation}
where $p(M(s) < n)$ is the probability of the state $s$ in training data after being pruned by State-DN that its $M(s)<n$. For example, if our predictor is perfect, $\alpha=1$, all the states with $M(s) \leq n$ will stop at $n$ simulations, while others will continue to finish $N_{max}$ simulations. 
\begin{lemma}
\label{lemma:decide_n}
The problem $\mathop{\arg\min}\limits_n f(\alpha, n)$ have the same solution for any $\alpha \in [0, 1]$.
\end{lemma}

We provide the detailed proof in the appendix. According to Lemma~\ref{lemma:decide_n}, we can safely select the cut point $n$ for MCTS-DN to decide whether the agent should keep searching or not by assuming it is a perfect predictor ($\alpha=1$), hence we have:
\begin{equation}
\label{equ:argmin_n}
    n=\mathop{\arg\min}\limits_{n'} (f(1,n')), 
\end{equation}
which can be solved exactly in a linear scan.  

Since training MCTS-UN needs MCTS features $T(s,n)$ of a state $s$ (root state of the search tree) after $n$ simulation ($n\leq N_{max}$) as input, we collect extra data while generating the training records of State-UN. The followings are the extra data we collected. More data can be stored for more complicated MCTS features $T(s,n)$.

For each state $s_0$ in training record, we collect a pair $(a_i, v_i)$ from each simulation, where $a_i$ is the first action on the selection path in the $i$th simulation, and $v_i$ is the evaluation result from the view of root state. For example, assume the root state $s_0$ is black's turn. If the selected leaf state $s_T$ is also black's turn, then $v_i = v(s_T)$; otherwise if $s_T$ is white's turn,  $v_i = -v(s_T)$. Hence, for each game state, we end up collecting a sequence of pairs $(a_2, v_2), (a_3, v_3), ..., (a_{N_{max}}, v_{N_{max}})$. Note that the first simulation evaluates the root state and does not select any action, hence there is no $(a_1, v_1)$. We also prune some of the training data that are certain in the beginning $U(s,1) = 0$ and can be identified by State-UN easily. This helps us solve the data imbalance issue for training MCTS-UN. 

With the extra data, we can generate MCTS features $T(s,n)$ of a state $s$ (root state of the search tree) after any $n$ simulation ($n\leq N_{max}$), which is described as followings:

MCTS features $T(s,n)$ is a tensor with dimensions of $channel\_num\times board\_size \times board\_size$, where our $channel\_num=7$. Each channel contains certain types of information about each action. Take Go as an example: channel 0 ($T(s,n)[0]$) stores the prior probability provided by the original PV-NN. Hence, $T(s,n)[0][1][2]$ stores the prior probability of playing at position $(1,2)$. Channel $1$ to $3$ store the information of the first layer of the search tree, including the policy ($N(s,a)/N(s)$) with $\tau=1$, the action value ($Q(a|s)$), and the standard deviation of action value. If an action $a'$ haven't been visited after $n$ simulations, then action value $Q(a|s)=-1$ and the standard deviation equals to $1$. Channel $4$ to $6$ generated by the latest half ($n/2$ to $n$) of the simulations to provide MCTS-UN the temporal information about the search, which also includes the policy, the action value, and the standard deviation of action value. For instance, the latest half policy can be calculated by $\pi(*|s,n) \times 2 - \pi(*|s,(n/2))$. The temporal information can show the trend of the search. For example, if the current best action is not selected in the last half of simulations, this may indicate that the MCTS is trying to switch to an alternative solution. Note that our MCTS features $T(s,n)$ do not have any information of $n$, since we hope to train our MCTS-UN on a single $n$ and use it on any others. If there are information of $n$ in $T(s,n)$, the features $T(s,n')$ will be out of the training distribution. For the same reason, $T(s,n)$ also does not have the information of $N_{max}$ since we also want to scale up to a larger $N_{max}$.

Finally, $\text{MCTS-UN}(s, T(s,n)) = (u,\mathbf{p},v)$ is trained as follows:
\begin{equation}
    \label{equ: mcts-un-loss}
    l =  (u - U(s,n))^2 + c_1 ((v-z)^2 - \pmb{\pi}^T log(\mathbf{p}))  + c_2\Vert{\theta}\Vert^2, 
\end{equation}
Same as State-UN, we also train it to learn the policy and value so that more information can be extracted. During training, we randomly dropout MCTS features or state features to ensure that MCTS-UN will not overly rely on any one of the features.

\subsection{Dynamic Simulation MCTS}
\label{sec:DSMCTS}
Although our predictors' accuracy can achieve more than 90\% accuracy, it is still non-trivial to use in practice since every wrong prediction can lead to losing directly. Moreover, due to the game's high complexity, it is possible to encounter many unseen data. Therefore, we need to be as careful as possible by making the accuracy of positive samples ($U(s,n)=1$) close to 100\%. Base on this idea, we propose Dynamic Simulation MCTS, which only stops the search when it is very certain. Otherwise, we will continue the search and recheck it at the next checking point. 

\begin{algorithm}[h]
\caption{Dynamic Simulation MCTS}
\label{alg:DS-MCTS}
\KwIn {State $s$, Maximum simulation $N_{max}$, Check points $\boldsymbol{c}$, Thresholds $\boldsymbol{thr}$}
\KwOut {policy $\pi$}
\LinesNumbered
\SetKwFunction{FStateUN}{state\_UN}
\SetKwFunction{FMCTSUN}{MCTS\_UN}
\SetKwFunction{FT}{T}
\SetKwFunction{Fsim}{MCTS\_simulation}
\SetKwFunction{FLen}{len}
    
\uIf{ \FStateUN$(s) < \boldsymbol{thr}[0]$}
    {\KwRet $p(*|s)$}
$i \gets 1$ 

\For{$n=1 : N_{max}$}{
    \uIf{$i< \FLen(\boldsymbol{c})$ and $n=\boldsymbol{c}[i]$}{
        \uIf{\FMCTSUN$(s, \FT(s,n)) < \boldsymbol{thr}[i]$}{
            \KwRet $\pi(*|s,n)$
        }
        $i \gets i + 1$
    }
    \Fsim$()$
}
\KwRet $\pi(*|s,N_{max})$
\end{algorithm}

Dynamic Simulation MCTS is shown in Algorithm~\ref{alg:DS-MCTS}. The input includes the given state $s$, maximum simulation count $N_{max}$, a list of checkpoints $\boldsymbol{c}$, and a list of thresholds $\boldsymbol{thr}$. Checkpoints list $\boldsymbol{c}$ is sorted and indicates when we should check our uncertainty. For example, the first checkpoint $\boldsymbol{c}[0]=0$ is at the beginning of the search.  The threshold list $\boldsymbol{thr}$ is the threshold for each checkpoint. 

At the beginning of the search, we obtain $u$ by the original PV-NN or the State-UN at line 1. If the output $u$ is $<\boldsymbol{thr}[0]$ (line 1), we will return the policy provided by PV-NN (line 2). Otherwise, the search starts (line 4). During the search, when the current simulation count $n$ reaches a checkpoint $\boldsymbol{c}[i]$ (line 5), we then evaluate the current uncertainty by using $\text{MCTS-UN}(s, T(s,\boldsymbol{c}[i]))$. If the output $u$ of MCTS-UN is smaller than $\boldsymbol{thr}[i]$ (line 6), then the search will stop (line 7). Otherwise, the search continues (line 9) and wait for the next checkpoint reaches (line 8). 

The number of checkpoints should not be too many since there is still cost for forwarding MCTS-UN. We normally selects the second checkpoint same as the training data $\boldsymbol{c}[1]=n$, where $n$ is selected by Equ.~\ref{equ:argmin_n}. For other checkpoints, we use $\boldsymbol{c}[i]=2 \times \boldsymbol{c}[i-1]$. Normally, when more simulations are executed, MCTS-UN's prediction will be more accurate. Therefore, we can use a larger threshold for a larger checkpoint, as long as most ($>96\%$) of the states that need to keep searching will be identified. This can be verified by evaluating the threshold on the self-play record first.

\section{Related Work}
Resource allocation~\cite{hyatt1984using,vsolak2009time,donninger1994recherche,markovitch1996learning} has a long history in game playing since the total thinking time is normally fixed (sudden death). However, it is not intensively studied in MCTS~\cite{huang2010time, baier2015time}, which is very different from the alpha-beta search because MCTS is an any-time algorithm.  PV-MCTS trained by AlphaZero is even more different from traditional searching since the action provided by PV-NN is already decent, and our model needs to evaluate the probability of other actions being better.

Resource allocation schemes can be divided into three categories: static, semi-static, and dynamic strategies based on decision time~\cite{markovitch1996learning}. Static strategies decide at the beginning of the game, and semi-static strategies decide before each search is started. These strategies usually focus on reserving time for the critical game stages. One of the semi-static strategies proposed by \citet{kocsis2000learning} uses a neural network to predict the search depth of the alpha-beta search. The search depth will follow the network's prediction. When the time runs out, the search depth will always be 1. Unlike state-UN, it is trained on a fixed total thinking time; hence we need to retrain the model when the total thinking time is changed. Also, since it is semi-static strategies, once the model predicts a search depth, it cannot change it according to the current searching result.

Our method belongs to dynamic strategies that allocate resources during the search. 
One of the most popular dynamic strategy "STOP"~\cite{baier2015time} terminates the search when the best move will not change even if the remaining simulations all select the current second-best move. Although "STOP" guarantees the same best move, it can only terminate the search after $N_{max}/2$ of simulations. ProbCut~\cite{buro1995probcut} can also be considered as dynamic strategies on a higher level. It speeds up alpha-beta search by pruning the search tree according to shallow subtree searching results. However, there is no obvious way to use it on MCTS, since MCTS has already use UCT~\cite{kocsis2006bandit} or PUCT~\cite{rosin2011multi} algorithm to handle subtrees that are not promising.  

Beyond tree searching, many methods have been proposed to quantify the uncertainty. Some methods utilize the original network's input and output~\cite{guo2017calibration, hendrycks2016baseline,liang2017enhancing},
some methods ensemble randomized models~\cite{gal2016dropout, lakshminarayanan2017simple}, some methods query model's understanding of the environment~\cite{pathak2017curiosity,burda2018exploration}, and some use Bayesian methods~\cite{guo2017calibration,azizzadenesheli2018efficient}. However, non of the above methods can utilize the information provided by the current tree search, since they can only provide uncertainty quantification for a static  model, while tree search is a dynamic process.
\section{Experiment}

\begin{figure*}[ht]
\minipage{0.3\textwidth}
        
\begin{tikzpicture}[scale=0.5]
\begin{axis} [xlabel = Average Simulation Ratio,
    ylabel = Win Rate,
    legend pos= south east,
    grid style=dashed,
    yscale=9/8,
    ymajorgrids=true,
    legend style={nodes={scale=0.8, transform shape}},
    xmin=0.24,
    xmax=0.61,
]

\addplot [
color=darkspringgreen!66!white,
thick,
mark options={fill=darkspringgreen, draw opacity=1, fill opacity=1},
mark=x,
line width=1.2pt,
] 
table[
col sep=comma, 
header=true, 
x=mp05_xr, 
y=mp05_y
] 
{experiment/mp/mp05.csv};
\addlegendentry{PV-NN~$_{\tau=0.5}$}


\addplot [
color=darkspringgreen,
thick,
mark=triangle*,
line width=1.2pt,
mark options={fill=darkspringgreen, draw opacity=0, fill opacity=1},
mark size=3pt,
dashed,
] 
table[
col sep=comma, 
header=true, 
x=mp1_xr, 
y=mp1_y,
] 
{experiment/mp/mp1.csv};
\addlegendentry{PV-NN~$_{\tau=1.0}$}


\addplot [
color=darkspringgreen!66!black,
thick,
mark=square,
line width=1.2pt,
] 
table[
col sep=comma, 
header=true, 
x=mp15_xr, 
y=mp15_y
] 
{experiment/mp/mp15.csv};
\addlegendentry{PV-NN~$_{\tau=1.5}$}

\addplot [
color=darkbrown,
thick,
mark=*,
line width=1.2pt,
mark options={fill=darkbrown, draw opacity=0},
] 
table[
col sep=comma, 
header=true, 
x=dnn80_xr, 
y=dnn80_y
] 
{experiment/dnn/dnn80.csv};
\addlegendentry{State-UN}

\addplot [
color=black,
solid,
line width=1.2pt,
]
coordinates{
 (400/1600,0.2955) (480/1600, 0.31)(560/1600,0.355)(640/1600,0.36)(720/1600,0.3705)(800/1600,0.41)(880/1600,0.4285)(960/1600,0.4355)
};
\addlegendentry{Baseline}

\end{axis}
\end{tikzpicture}
    \captionof{figure}{The results of using the PV-NN and State-UN.}
    \label{fig:stop_at_beginning}
\endminipage\hfill
\minipage{0.3\textwidth}
            \begin{tikzpicture}[scale=0.5]
        \begin{axis} [   xlabel = Average Simulation Ratio,
            ylabel = Win Rate,
            legend pos= south east,
            grid style=dashed,
            yscale=9/8,
            ymajorgrids=true,
            legend style={nodes={scale=0.8, transform shape}},
            xmin=0.24,
            xmax=0.61,
        ]

        \addplot [
        color=carnelian!44,
        thick,
        mark=10-pointed star,
        line width=1.2pt,
        mark options={fill=carnelian!44, fill opacity=1},
        mark size=3pt,
        ] 
        table[
        col sep=comma, 
        header=true, 
        x=s160_xr, 
        y=s160_y
        ] 
        {experiment/cp/s160.csv};
        \addlegendentry{MCTS-UN~$_{\boldsymbol{c}[1]=160}$}
        
        \addplot [
        color=carnelian!77,
        thick,
        mark=pentagon,
        mark options={solid, fill=carnelian!77, fill opacity=1},
        dashed,
        line width=1.2pt,
        ] 
        table[
        col sep=comma, 
        header=true, 
        x=s400_xr, 
        y=s400_y
        ] 
        {experiment/cp/s400.csv};
        \addlegendentry{MCTS-UN~$_{\boldsymbol{c}[1]=400}$}

        \addplot [
        color=carnelian,
        mark=diamond*,
        mark size=2.3pt,
        line width=1.1pt,
        mark options={fill=carnelian, fill opacity=1},
        ] 
        table[
        col sep=comma, 
        header=true, 
        x=s800_xr, 
        y=s800_y
        ] 
        {experiment/cp/s800.csv};
        \addlegendentry{MCTS-UN~$_{\boldsymbol{c}[1]=800}$}
        
        \addplot [
        color=darkbrown,
        thick,
        mark=*,
        line width=1.2pt,
        mark options={fill=darkbrown, draw opacity=0},
        ]
        table[
        col sep=comma, 
        header=true, 
        x=dnn80_xr, 
        y=dnn80_y
        ] 
        {experiment/dnn/dnn80.csv};
        \addlegendentry{State-UN}
        
        \addplot [
        color=black,
        solid,
        line width=1.2pt,
        ]
        coordinates{
         (400/1600,0.2955)(480/1600,0.31)(560/1600,0.355)(640/1600,0.36) (720/1600,0.3705)(800/1600,0.41)(880/1600,0.4285)(960/1600,0.4355)
        };
        \addlegendentry{Baseline}

        \end{axis}
        \end{tikzpicture}
    \captionof{figure}{The MCTS-UN results using different check point $\boldsymbol{c}[1]$.}
    \label{fig:MCTS_UN}
\endminipage\hfill
\minipage{0.3\textwidth}%
          \begin{tikzpicture}[scale=0.5]
        \begin{axis} [   xlabel = Average Simulation Ratio,
            ylabel = Win Rate,
            legend pos= south east,
            grid style=dashed,
            yscale=9/8,
            ymajorgrids=true,
            legend style={nodes={scale=0.8, transform shape}},
            xmin=0.24,
            xmax=0.61,
            ymin=0.27,
        ]

        \addplot [
        color=carnelian!44,
        thick,
        mark=10-pointed star,
        line width=1.2pt,
        mark options={fill=carnelian!44, fill opacity=1},
        mark size=3pt,
        ] 
        table[
        col sep=comma, 
        header=true, 
        x=s160_acc_xr, 
        y=s160_acc_y
        ] 
        {experiment/cp/s160_acc.csv};
        \addlegendentry{MCTS-UN~$_{\boldsymbol{c}[1]=160}$}
        
        \addplot [
        color=carnelian!77,
        thick,
        mark=pentagon,
        mark options={solid, fill=carnelian!77, fill opacity=1},
        dashed,
        line width=1.2pt,
        ] 
        table[
        col sep=comma, 
        header=true, 
        x=s400_acc_xr, 
        y=s400_acc_y
        ] 
        {experiment/cp/s400_acc.csv};
        \addlegendentry{MCTS-UN~$_{\boldsymbol{c}[1]=400}$}

        \addplot [
        color=carnelian,
        mark=diamond*,
        mark size=2.3pt,
        line width=1.1pt,
        mark options={fill=carnelian, fill opacity=1},
        ] 
        table[
        col sep=comma, 
        header=true, 
        x=s800_acc_xr, 
        y=s800_acc_y
        ] 
        {experiment/cp/s800_acc.csv};
        \addlegendentry{MCTS-UN~$_{\boldsymbol{c}[1]=800}$}
        
        \addplot [
        color=darkbrown,
        thick,
        mark=*,
        line width=1.2pt,
        mark options={fill=darkbrown, draw opacity=0},
        ] 
        table[
        col sep=comma, 
        header=true, 
        x=dnn80_acc_xr, 
        y=dnn80_acc_y
        ] 
        {experiment/dnn/dnn80_acc.csv};
        \addlegendentry{State-UN}

        \addplot [
        color=black,
        solid,
        line width=1.2pt,
        ]
        coordinates{
          (425/1600,0.3375)(495/1600,0.35595)(565/1600, 0.36375)(630/1600,0.399) (699/1600,0.4045)(765/1600,0.41185) (834/1600,0.453) (964/1600,0.48825)
        };
        \addlegendentry{Baseline}

        \end{axis}
        \end{tikzpicture}
  \captionof{figure}{Applying "STOP" algorithm to MCTS-UN, State-UN, and Baseline.}
  \label{fig:MCTS_UN_STOP}
\endminipage
\end{figure*}




In this section, we empirically investigate our method mainly on 9x9 NoGo and some on 9x9 Go. 

\subsubsection{Experiment setting for NoGo} 
NoGo is one of the tournament items in the Computer Olympiad \cite{nogo}. We choose NoGo in our experiments since the searching space is relatively low compared with 19x19 Go while maintaining many similar characteristics~\cite{chou2011revisiting}.

The NoGo agent we used has a 98\% win rate against HaHaNoGo \cite{HaHaNoGo}, which won the champion on the 2016 TAAI competition. The PV-NN is a ResNet~\cite{he2016deep} with ten blocks and 128 filters. State-UN and MCTS-UN have ten blocks and 196 filters and are trained by 20,000 self-play games with $N_{max}=1600$. MCTS-UN is trained by $n=160$, where $n$ is acquired by Equation~\ref{equ:argmin_n}. Each experiment played at least 2,048 games; therefore, the standard deviation was less than 1.1\%. The default opponent is PV-MCTS with $opp\_sim = 1600$ simulations on each move. We do not other programs because their are much weaker ($<10\%$ win rate) than our agent that is trained by AlphaZero. The default $N_{max}$ for DS-MCTS method is also 1600. 

\subsubsection{Performance of original PV-NN and State-UN}
First, we investigated the performance of DS-MCTS that only checks at the beginning of the search using PV-NN or State-UN. The results are shown in Fig.~\ref{fig:stop_at_beginning}. We use average simulation ratios as the x-axis to show the acceleration, namely, if the average simulation is $E[n]$, then the average simulation ratio is $E[n]/opp\_sim$. The baseline (black line in Fig.~\ref{fig:stop_at_beginning}) is drawn by PV-MCTS with different simulations. We have also tested stopping the searching randomly \cite{wu2019accelerating}. The results show that it is much weaker than our baseline. 

We experiment using Equ.~\ref{equ:CF} with three different temperatures ($\tau=\{0.5,1,1.5\}$). Since using different thresholds $\boldsymbol{thr}[0]$ will result in different average simulations, we properly select the set of thresholds (see appendix) for each temperature so that they can be compared under similar average simulations. The result is shown as the green lines in Fig.~\ref{fig:stop_at_beginning}, where each data point is generated by a threshold $\boldsymbol{thr}[0]$ in the set. The results show that PV-NN can achieve more than $45\%$ win rate against the original PV-MCTS with only using less than $60\%$ of time. Temperature $\tau = 1$ even achieves a $48.6\%$ win rate with an average of $925$ simulations, where the original PV-MCTS (black) can only achieve about $42\%$ win rate using the same simulations. However, as the threshold $\boldsymbol{thr}[0]$ increase (average simulation decrease), the win rates drop quickly. It even performs worse than the baseline when the average simulation ratio is smaller than 0.4. This implies that some critic states that require searching are misclassified, and the game will be unsavable even if searching longer on other states. 

Same as original PV-NN, we also use a set of $\boldsymbol{thr}[0] = {0.05, 0.1, 0.2, ..., 0.7}$ to show how the win rate decrease when more states play without search according to the State-UN. The brown line in Fig.~\ref{fig:stop_at_beginning} shows that the uncertainty prediction $u$ provided by State-UN is much better than using the original PV-NN. Especially when using less than 40\% of the resource, State-UN can still have a 43\% win rate. This also indicates that our definition of uncertainty $U(s,n)$ for MCTS tree search serves properly on training State-UN. Since $\boldsymbol{thr}[0]=0.1$ achieve a 49\% win rate with only 801 average simulations, we will use it for following experiments.

\setlength{\tabcolsep}{8pt}

\begin{table*}[t]
\footnotesize
  \caption{The results of testing on larger simulations with $\boldsymbol{thr}[0]=0.1$}
  \label{table: large sim}
  \centering
  
  \begin{tabular}{lccccccccc}

    \toprule
    & $\boldsymbol{c}[i]$ & -- & 160 & 400 & 800 & 1600 & 800/1600 & 400/800/1600 & 160/400/800/1600\\
    \cmidrule(r){2-10}
    
    & $\boldsymbol{thr}[i]$ & -- &  .025 & .025 & .05 & .05 & .05/.05 & .025/.05/.05 & .025/.025/.05/.05 \\
    \midrule
    
    \multirow{2}{*}[-0.7ex]{4000 vs 1600} 
    
    &  {Avg Sim} & 2226 & 1922 & 1858 & \textbf{1584} & 1748 & \textbf{1460} & {1401} & 1351\\
    
    \cmidrule(r){2-10}
    
    & {Win Rate} & 62.9\% & 59.7\% & {61.8\%} & \textbf{61.4\%} & 60.9\% & \textbf{61.5\%} & {59.1\%} & 54.8\%\\

    \cmidrule(r){1-10}

    \multirow{2}{*}[-0.7ex]{6400 vs 6400} 
    
    & {Avg Sim} & 3914 & 3341 & 3273 & 2815 & 2983 & 2539 & \textbf{2453} & 2364\\
    
    \cmidrule(r){2-10}
    
    & {Win Rate} & {48.0\%} & 44.1\% & {46.9\%} & 46.1\% & 46.5\% & 46.5\% & \textbf{46.5\%} & 44.7\%\\
    
    \bottomrule
  \end{tabular}
\end{table*}

\subsubsection{Performance of MCTS-UN}

Although our MCTS-UN is only trained with tree information $T(s, 160)$ at 160 simulations, we want to show that it can be used on other simulations such as 400 and 800. Hence, in these experiments, we add only a checkpoint on different simulations ($\boldsymbol{c}[1]={160,400,800}$) of the searching and use a set of thresholds ($\boldsymbol{thr}[1]$) to show the trade-off. The State-UN is used in the beginning with a threshold $\boldsymbol{thr}[0] = 0.1$.  Fig.~\ref{fig:MCTS_UN} shows that adding a single checkpoint $\boldsymbol{c}[1]$ during the search can achieve a higher performance than only using State-UN, even if the checkpoint is different from the simulations MCTS-UN is trained. For example, with $\boldsymbol{c}[1]=160, \boldsymbol{thr}[1]=0.025$ we have $49.2\%$ win rate with 715 average simulations and with setting of $\boldsymbol{c}[1]=800, \boldsymbol{thr}[1]=0.2$ we have $48.5\%$ win rate with 591 average simulations. Note that $\boldsymbol{c}=800$ has a higher win rate when having lower average simulations. This shows that even if MCTS-UN does not check and stop the searching until 800 simulations, the quality of MCTS-UN with the tree information $T(s,800)$ allows us to stop the searching of more states (using a higher $\boldsymbol{thr}[1]$) without dropping the win rate. Hence, using only one checkpoint $n=160$, which was selected by Equ.~\ref{equ:argmin_n}, is an efficient way to train MCTS-UN.

\subsubsection{Combining with the STOP strategy}

Next, we apply the "STOP" algorithm \cite{baier2015time}, as mention in the related work section, to both our methods and baseline programs. The results in Fig.~\ref{fig:MCTS_UN_STOP} shows that using "STOP" can further accelerate the search since all lines (including baseline) move slightly to the left comparing to Fig.~\ref{fig:MCTS_UN}. For example, when $\boldsymbol{c}[1]=800, \boldsymbol{thr}[1]=0.2$ we achieves 48.5\% win rate with only 555 simulations. The results also show that our methods still perform better than the baseline. For instance, to achieve more than 48\% win rate, PV-MCTS still needs 964 simulations even with the "STOP" algorithm. Note that we can compare our method with "STOP" directly by comparing the brown line and red line in Fig.~\ref{fig:MCTS_UN} and the black line in Fig.~\ref{fig:MCTS_UN_STOP}. Our method has a better result than "STOP", even if only using State-UN.

\subsubsection{Testing on larger simulations}
In this experiment, we want to show that although our MCTS-UN is trained by the data generated with $N_{max} = 1600$, it can be used in a larger $N'_{max}$. Moreover,  with more checkpoints, our method can perform even better. 

In the first experiment (4000vs1600), we increase the $N_{max}$ of DS-MCTS to 4000 and still fight against PV-MCTS with 1600 simulations per move. Besides showing our method can scale up, we also want to show that our method has a higher win rate under a similar average simulation count. Table~\ref{table: large sim} shows the results. The first column only uses State-UN with $\boldsymbol{thr}[0]=.1$. Columns from 6th to 8th show the results that use MCTS-UN more than one time. For example, the last column used checkpoints $\boldsymbol{c}=\{0,160,400,800,1600\}$ with thresholds $\boldsymbol{thr}=\{.1,.025,.025,.05,.05\}$ respectively. The result of (800, .05) shows that we can have a 61.4\% win rate against PV-MCTS that uses a similar simulation count. The result (800/1600, .05/.05) shows that we can reduce the simulation count without decreasing the win rate by using MCTS-UN to check on both 800 and 1600 simulations.

In the second experiment (6400vs6400), we increase the $N_{max}$ of DS-MCTS to 6400 and fight against PV-MCTS with 6400 simulations. 
We want to test if our method can scale up and fight against a much different and stronger agent. The last row of Table~\ref{table: large sim} shows we achieve more than a 46\% win rate against 6400 simulations with most of the settings. Moreover, with the setting of $\boldsymbol{c}=\{0,400,800,1600\}$, we can have a 46.5\% win rate against 6400 simulations with only 2453 average simulations. Note that the win rate of (2400vs6400) for PV-MCTS is only 34\%. Thus we show that MCTS-UN can be used when both $N'_{max}$ is much larger, and the opponent is much stronger than training data. 

\subsubsection{Experiments on 9x9 Go} 
We further apply our methods to 9x9 Go to demonstrate it is general to other games.
The PV-NN of our AlphaZero 9x9 go agent is ResNet with 20 blocks and 256 filters. The training data is generated in the same way as NoGo. 

There are two differences between the experiments of Go and NoGo. First, for each game, instead of starting from the empty position, we randomly select an opening from a public 9x9 Go opening book~\cite{goopening} as the starting position. Our State-UN and MCTS-UN are trained using self-play records. Using openings can help us verify whether our models can handle unseen positions in the training data. 
Second, for simplicity, we choose $0.1$ as the threshold for every checkpoint.

The first experiment is DS-MCTS vs. PV-MCTS with both 1600 simulations. With only State-UN, the average simulation ratio can drop to 79\% with about 50\% win rate. Adding a checkpoint at ${160,400,800}$ separately can reduce the average simulation ratio to ${67.6\%, 67.5\%, 71.1\%}$ and remain more than 46.5\% win rate. If we use three checkpoints together, the average simulation ratio can be further reduced to $59.4\%$. We then test this setting with $N_{max}=3200$ against PV-NN with 3200 simulations per move and still get a $44.9\%$ win rate. This shows that our method can scale up to a larger $N_{max}$, even if the position is unseen and the opponent is stronger. Other full results of 9x9 Go are shown in the appendix.

\section{Conclusion and Future Work}
In this paper, we first define the uncertainty for tree searching. We then train State-UN and MCTS-UN to predict the uncertainty. We also propose Dynamic Simulation MCTS to utilize those networks. Experimental results on NoGo and Go show that DS-MCTS can speed up the inference time and maintain the win rate. Moreover, although training on a small maximum simulation, MCTS-UN can scale up to a much larger one. After fine-tuning the thresholds on NoGo, DS-MCTS can speed up more than 2.5x and still maintain a 46.5\% win rate under a much larger $N_{max}$.

One of our future work is applying our method for generating self-play records during AlphaZero training. \citet{wu2019accelerating} shows that randomly reduce the simulation count on 75\% of states is still able to train a strong AI. Hence, we believe that we can reduce simulation on even more states if we select the states properly instead of random. In our inner experiments of using PV-NN to estimate uncertainty, 19x19 Go is much easier to speed up than 9x9 Go.

Another future work is inspired by ProbCut~\cite{buro1995probcut}. We can stop not only the searching of the root state but also any state in the search tree. Also, to improve MCTS-UN, we need to input more information of the search tree. A promising way is using graph neural networks~\cite{wu2020comprehensive} to gather deeper information from the search tree.

\section{Acknowledgments}
This research is partially supported by the Ministry of Science and Technology (MOST) under Grant Number MOST 109-2634-F-009-019 through Pervasive Artiﬁcial Intelligence Research (PAIR) Labs. The computing resource is partially supported by national center for high-performance computing (NCHC).

\bibliography{aaai21}
\appendix
\section*{Appendix}

\setlength{\tabcolsep}{10pt}

\begin{table*}[htb]
\footnotesize
  \caption{The results of Go with $\boldsymbol{thr}[0]=0.1$}
  \label{table: large sim}
  \centering
  
  \begin{tabular}{lccccccc}

    \toprule
    & $\boldsymbol{c}[i]$ & -- & 160 & 400 & 800 & 400/800 & 160/400/800\\
    
    & $\boldsymbol{thr}[i]$ & -- &  .1 & .1 & .1 & .1/.1 & .1/.1/.1 \\
    \midrule
    
    \multirow{2}{*}[-0.7ex]{1600 vs 1600} 
    
    &  {Avg Sim Ratio} & 79.2\% & 67.6\% & 67.5\% & 71.1\% & 64.0\% & 59.4\%\\
    

    & {Win Rate} & 47.9\% & 46.4\% & 46.5\% & 48.0\% & 45.0\% & 45.3\% \\

     \midrule

    \multirow{2}{*}[-0.7ex]{3200 vs 3200} 
    
    & {Avg Sim Ratio} & 81.98\% & 67.1\% & 65.7\% & 65.5\% & 60.9\% & 58.0\% \\
    
    
    & {Win Rate} & {47.3\%} & 44.4\% & {44.8\%} & 46.7\% & 44.7\% & 44.9\% \\
    
    \bottomrule
  \end{tabular}
\end{table*}

\section{Proof of Lemma 3.1}
\begin{lemma}
\label{lemma:decideL}
The problem $\mathop{\arg\min}\limits_n f(\alpha, n)$ have the same solution for any $\alpha \in [0, 1]$. 
\end{lemma}
where $f(\alpha, n)$ is:
\begin{equation}
\begin{split}
     f(\alpha,n)=
     (n \times \alpha + N_{max} \times (1-\alpha)) p(M(s) < n)\\
     + N_{max}p(M(s)\geq n)
\end{split}
\end{equation}

\begin{proof}
Assume $f(1, n_1) < f(1, n_2)$, we have
\begin{equation}
\begin{split}
    n_1\times p(M(s) < n_1) + N_{max}p(M(s)\geq n_1)\\
    < n_2\times p(M(s) < n_2) + N_{max}p(M(s)\geq n_2)
\end{split}
\end{equation}
Times $\alpha$ at both sides:
\begin{equation}
\begin{split}
    \alpha \times  n_1\times p(M(s) < n_1) + \alpha \times N_{max}p(M(s)\geq n_1)\\
    < \alpha \times n_2\times p(M(s) < n_2) + \alpha \times N_{max}p(M(s)\geq n_2)
\end{split}
\end{equation}
Add $(1-\alpha) \times N_{max}$ to both sides:
\begin{equation}
\begin{split}
    (\alpha n_1   + N_{max} (1-\alpha)) p(M(s) < n_1)
     \\ 
     + N_{max}p(M(s)\geq n_1)\\
     < ( \alpha n_2  + N_{max} (1-\alpha)) p(M(s) < n_2)
     \\
     + N_{max}p(M(s)\geq n_2)
\end{split}
\end{equation}

Hence we have $f(\alpha, n_1) < f(\alpha, n_2)$. Therefore, for every $\alpha$, $\mathop{\arg\min}\limits_n f(\alpha, n)$ is same as $\mathop{\arg\min}\limits_l f(1, l)$.  
\end{proof}

\section{Experiment Settings of NoGo}

Although our goal seems simple, training MCTS-UN is still hard due to the data imbalance. Only less than 10\% of states that are $M(s) > 160$. To train MCTS-UN, we first generate 20,000 self-play records using 800 simulations. We then re-evaluate the states in the records using $N_{max}=1,600$ simulations to calculate the $M(s)$ of each state. We set $c_1=0.1$ for both loss functions. After training the State-UN, we use a threshold of $0.05$ to prune out about $1/3$ of the data with $M(s)=1$, since they are easy to be identified by the State-UN. During training MCTS-UN, we also increase the probability of sampling states that are $M(s) > 160$, in order to handle data imbalance problems. Each state sampled has 50\% to be trained with both state and MCTS features; it has 25\% to be trained with only state features, and the last 25\% to be trained with only MCTS features. In this way, the model will not overly rely on one of the features to determine uncertainty.

In fig. 3, we use different sets of thresholds for different temperatures $\tau$, so that their result can be compared under the same simulation ratio. We select the thresholds according to the distribution of training data. For $\tau=0.5$, the thresholds are $\{0.094,0.137,0.23,0.287,0.332\}$. For $\tau=1$, the thresholds are $\{0.35,0.387,0.495,0.527,0.563\}$. For $\tau=1.5$, the thresholds are $\{0.567,0.605,0.655,0.681,0.706\}$.

\section{Experiment of Go}

Table~\ref{table: large sim} shows the full results of experiments on Go. All the thresholds are set to 0.1 without any fine-tuned. The results show that when we use a larger simulation, the are more room for us to speed up the agent and still remain almost the same win rate, especially for the larger checkpoints.

\end{document}